# Optimizing F-Measures: A Tale of Two Approaches

**Nan Ye**                                                                                    YENAN@COMP.NUS.EDU.SG
Department of Computer Science, National University of Singapore, Singapore 117417

**Kian Ming A. Chai**                                                                          CKIANMIN@DSO.ORG.SG
DSO National Laboratories, Singapore 118230

**Wee Sun Lee**                                                                               LEEWS@COMP.NUS.EDU.SG
Department of Computer Science, National University of Singapore, Singapore 117417

**Hai Leong Chieu**                                                                           CHAILEON@DSO.ORG.SG
DSO National Laboratories, Singapore 118230

## Abstract

F-measures are popular performance metrics, particularly for tasks with imbalanced data sets. Algorithms for learning to maximize F-measures follow two approaches: the empirical utility maximization (EUM) approach learns a classifier having optimal performance on training data, while the decision-theoretic approach learns a probabilistic model and then predicts labels with maximum expected F-measure. In this paper, we investigate the theoretical justifications and connections for these two approaches, and we study the conditions under which one approach is preferable to the other using synthetic and real datasets. Given accurate models, our results suggest that the two approaches are asymptotically equivalent given large training and test sets. Nevertheless, empirically, the EUM approach appears to be more robust against model misspecification, and given a good model, the decision-theoretic approach appears to be better for handling rare classes and a common domain adaptation scenario.

## 1. Introduction

F-measures (van Rijsbergen, 1974) or F-scores have been commonly used in tasks in which it is important to retrieve elements belonging to a particular class correctly without including too many elements of other classes. F-measures are usually preferred to accuracies as standard performance measures in information retrieval (Manning et al., 2008), particularly, when relevant items are rare. They are also popular in information extraction tasks such as named entity recognition (Tjong Kim Sang & De Meulder, 2003) where most of the elements do not belong to a named class.

Various methods have been proposed for optimizing F-measures. They fall into two paradigms. The empirical utility maximization (EUM) approach learns a classifier having optimal F-measure on the training data. Optimizing the F-measure directly is often difficult as the F-measure is non-convex. Thus approximation methods are often used instead. Joachims (2005) gave an efficient algorithm for maximizing a convex lower bound of F-measures for support vector machines, and showed it worked well on text classification. Jansche (2005) gave an efficient algorithm to maximize a non-convex approximation to F-measures using logistic regression models, and showed it works well on a text summarization problem. A simpler method is to optimize the F-measure in two stages: First learn a score function using standard methods such as logistic regression or support vector machines, then select a threshold for the score function to maximize the empirical F-measure. Though simple, this method has been found to be effective and is commonly applied, for example, in text categorization (Yang, 2001).

The decision-theoretic approach (DTA), advocated by Lewis (1995), estimates a probability model first, and then computes the optimal predictions (in the sense of having highest expected F-measure) according to the model. This method has not been commonly applied for F-measures, possibly due to the high computa-

Optimizing F-Measures: A Tale of Two ApproachesOptimizing F-Measures: A Tale of Two Approaches

tional complexity of existing algorithms for the prediction step. Assuming the independence of labels, Lewis showed that, in the optimal prediction, the probabilities of being positive for irrelevant items are not more than those for relevant items. He also gave a bound for expected F-measures, which can be computed in $O(n)$ time, but can be very loose. Based on Lewis's characterization, Chai (2005) gave an $O(n^3)$ time algorithm to compute optimal predictions, and he gave empirical demonstration for the effectiveness of DTA. Apparently unaware of Chai's work, Jansche (2007) solved the same problem in $O(n^4)$ time. For the general case when the labels are not necessarily independent, Dembczynski et al. (2011) gave an $O(n^3)$ time algorithm given $n^2+1$ parameters of the label distribution, but the parameters can be expensive to compute. They also showed that the independence assumption can lead to bad performance in the worst case, but on the practical datasets used in their experiments, methods assuming the independence assumption are at least as good as those not assuming independence.

We have only discussed works on binary classification. There are also algorithms for optimizing F-measures for tasks with structured output (Tsochantaridis et al., 2005; Suzuki et al., 2006; Daumé et al., 2009) and multilabel tasks (Fan & Lin, 2007; Zhang et al., 2010; Petterson & Caetano, 2010).

Optimality in EUM and DTA are different. EUM considers only instance classifiers (functions mapping instances to labels), and roughly speaking, an optimal classifier is an instance classifier having highest F-measure on a very large test set among all instance classifiers. On the other hand, DTA considers set classifiers (functions mapping sets of instances to sets of labels), and an optimal classifier in DTA is a set classifier having maximum expected F-measure among all set classifiers. Optimality in these two approaches are also achieved differently using different learning objectives. Unless otherwise stated, optimal classifiers refer to EUM-optimal classifiers, and optimal predictions refer to predictions by DTA-optimal classifiers.

In this paper, we study the relative effectiveness of the two approaches, and develop theories and algorithms for this purpose. We focus on binary classification, assuming the data is independently and identically distributed (i.i.d.). The contributions of this paper are as follows. In Section 2, we establish a consistency result for empirical maximization of F-measures, together with bounds on the rate of convergence. This provides some insights into the factors affecting the convergence rate in EUM. In particular, our bounds suggest that rare classes require more data for performance guarantee, which is consistent with our intuition. We then show that thresholding the true conditional distribution on a large i.i.d. test set can perform as well as the best instance classifier, justifying the popular hybrid approach of learning a conditional distribution followed by learning a threshold. We also show that an EUM-optimal classifier and a DTA-optimal classifier are asymptotically equivalent if the probability measure for any set of instances with the same conditional probability of being relevant is negligible.

In Section 3, we give a new $O(n^2)$ time algorithm for computing optimal predictions, assuming independence of labels. Our algorithm can compute optimal predictions on tens of thousand instances within seconds, significantly faster than previous algorithms which require hours or more. [1]

In Section 4, we compare EUM and DTA on synthetic and real datasets. Our theoretical results are useful in explaining the experimental results. Empirically, EUM seems more robust against model misspecification, but given a good model, DTA seems better for handling rare classes on small datasets and a common scenario of domain adaptation.

## 2. Theoretical Analysis

Let $X$ and $Y$ denote the input and output random variables. We assume there is a fixed but unknown distribution $P(X,Y)$ that generates i.i.d. $(X,Y)$ pairs during training and testing. We use $X$ and $Y$ to denote their domains as well. In this paper, $Y = \{0,1\}$, with 0 for the negative or irrelevant class and 1 for the positive or relevant class. $\mathrm{I}(\cdot)$ is the indicator function.

Let $D_n = \{(x_1, y_1), \ldots, (x_n, y_n)\}$ be a set of $n$ (possibly non-i.i.d.) examples, and let $\mathbf{x}$ and $\mathbf{y}$ denote $(x_1, \ldots, x_n)$ and $(y_1, \ldots, y_n)$ respectively. If the predicted labels are $\mathbf{s} = (s_1, \ldots, s_n)$, then *precision* $p(\mathbf{s}, \mathbf{y})$ is the number of true positives over the number of predicted positives, and *recall* $r(\mathbf{s}, \mathbf{y})$ is the number of true positives over the number of positives. $F_\beta$-*measure* (van Rijsbergen, 1974) $F_\beta(\mathbf{s}, \mathbf{y})$ is a weighted harmonic mean of precision and recall. Formally,

$$F_\beta(\mathbf{s}, \mathbf{y}) = \frac{(1+\beta^2)\sum_i s_i y_i}{\beta^2 \sum_i y_i + \sum_i s_i}, \qquad (1)$$

$p(\mathbf{s}, \mathbf{y}) = \sum_i s_i y_i / \sum_i s_i$ and $r(\mathbf{s}, \mathbf{y}) = \sum_i s_i y_i / \sum_i y_i$. Thus, $F_\beta = (1+\beta^2)/(\beta^2/r + p)$. In addition, $F_0$ is the precision and $F_\infty$ is the recall. $F_1$ is most frequently used in practice. Henceforth, we assume $\beta \in (0, \infty)$.

---

[1] See http://www.comp.nus.edu.sg/~yenan/.

### 2.1. Uniform Convergence and Consistency for EUM

Consider an arbitrary classifier $\theta : X \mapsto Y$. Let $F_{\beta,n}(\theta)$ denote the $F_\beta$ score of $\theta$ on $D_n$. Let $p_{ij,n}(\theta)$ be the empirical probability that a class $i$ instance is observed and predicted as class $j$ by $\theta$; that is, $p_{ij,n}(\theta) = \sum_{k=1}^n \mathrm{I}(y_k = i \wedge \theta(x_k) = j)/n$. Then

$$F_{\beta,n}(\theta) = \frac{(1+\beta^2)p_{11,n}(\theta)}{\beta^2(p_{11,n}(\theta) + p_{10,n}(\theta)) + p_{11,n}(\theta) + p_{01,n}(\theta)}$$

Let $p_{ij}(\theta) = \mathrm{E}(\mathrm{I}(Y = i \wedge \theta(X) = j))$, that is, the probability that a class $i$ instance is predicted as class $j$ by $\theta$. Under the i.i.d. assumption, for large i.i.d. sample, the law of large numbers implies that $p_{ij,n}(\theta)$'s converge to $p_{ij}(\theta)$'s. Thus $F_{\beta,n}(\theta)$ is expected to converge to

$$F_\beta(\theta) = \frac{(1+\beta^2)p_{11}(\theta)}{\beta^2\pi_1 + p_{11}(\theta) + p_{01}(\theta)}, \quad (2)$$

where $\pi_Y$ denotes $P(Y)$. Hence we can define this to be the $F_\beta$-measure of the classifier $\theta$. The above heuristic argument is formalized below. We often omit $\theta$ from the notations whenever there is no ambiguity. All proofs are in the supplement (See foonote 1).

**Lemma 1.** *For any $\epsilon > 0$, $\lim_{n\to\infty} \mathrm{P}(|F_{\beta,n}(\theta) - F_\beta(\theta)| < \epsilon) = 1$.*

By using a concentration inequality, such as the Hoeffding's inequality, in place of the law of large numbers, we can obtain a bound on the convergence rate.

**Lemma 2.** *Let $r(n,\eta) = \sqrt{\frac{1}{2n}\ln\frac{6}{\eta}}$. When $r(n,\eta) < \frac{\beta^2\pi_1}{2(1+\beta^2)}$, then with probability at least $1-\eta$, $|F_{\beta,n}(\theta) - F_\beta(\theta)| < \frac{3(1+\beta^2)r(n,\eta)}{\beta^2\pi_1 - 2(1+\beta^2)r(n,\eta)}$.*

We now show that training to maximize the empirical $F_\beta$ is consistent, using VC-dimension (Vapnik, 1995) to quantify the complexity of the classifier class.

**Theorem 3.** *Let $\Theta \subseteq X \mapsto Y$, $d = VC(\Theta)$, $\theta^* = \arg\max_{\theta\in\Theta} F_\beta(\theta)$, and $\theta_n = \arg\max_{\theta\in\Theta} F_{\beta,n}(\theta)$. Let $\bar{r}(n,\eta) = \sqrt{\frac{1}{n}(\ln\frac{12}{\eta} + d\ln\frac{2en}{d})}$. If $n$ is such that $\bar{r}(n,\eta) < \frac{\beta^2\pi_1}{2(1+\beta^2)}$, then with probability at least $1-\eta$, $F_\beta(\theta_n) > F_\beta(\theta^*) - \frac{6(1+\beta^2)\bar{r}(n,\eta)}{\beta^2\pi_1 - 2(1+\beta^2)\bar{r}(n,\eta)}$.*

The above bound indicates that for smaller $\pi_1$ and $\beta$, more samples are probably required for convergence to start occurring. When $r(n,\eta) < \frac{\beta^2\pi_1}{4(1+\beta^2)}$, the difference between $F_{\beta,n}(\theta)$ and $F_\beta(\theta)$ is at most $\frac{6(1+\beta^2)}{\beta^2\pi_1}r(n,\eta)$.

### 2.2. Optimality of Thresholding in EUM

We now consider a common EUM approach: learning a score function and then using a fixed threshold on the score function. This threshold is obtained by optimizing the F-measure on the training data.

Assume we know the true conditional distribution $P(Y|X)$. Consider the class $\mathcal{T}$ of probability thresholding classifiers of the form $\mathrm{I}_\delta(x) = \mathrm{I}(P(1|x) > \delta)$, and the class $\mathcal{T}'$ containing $\mathrm{I}'_\delta(x) = \mathrm{I}(P(1|x) \geq \delta)$. [2] $\mathcal{T} \cup \mathcal{T}'$ has VC dimension 1, so empirical maximization of F-measure for this class is consistent. Although $\mathcal{T} \cup \mathcal{T}'$ does not contain all possible classifiers on $X$, an optimal classifier can be found in this class. Let $t^* = \arg\max_{h\in\mathcal{T}\cup\mathcal{T}'} F_\beta(h)$.

**Theorem 4.** *For any classifier $\theta$, $F_\beta(\theta) \leq F_\beta(t^*)$.*

Thresholding is often applied on a score function $f : X \mapsto \mathbf{R}$, rather than on the true conditional distribution. For example, output of a support vector machine is commonly thresholded. Let $f_\delta(x) = \mathrm{I}(f(x) > \delta)$ and $f'_\delta(x) = \mathrm{I}(f(x) \geq \delta)$. Function $f$ is called an *optimal score function* if there is a $\delta$ such that $F_\beta(f'_\delta) = F_\beta(t^*)$. We give a sufficient condition for a score function to be optimal. A score function $f$ is *rank-preserving* if it satisfies $f(x_1) > f(x_2)$ iff $P(1|x_1) > P(1|x_2)$ for all $x_1, x_2 \in X$. The sufficient condition relates rank-preservation to optimality:

**Theorem 5.** *A rank-preserving function is an optimal score function.*

By Theorem 5, we can sidestep learning the true distribution and instead try to learn a function which is likely to be rank-preserving. An optimal score function may not be rank-preserving. For example, we can swap the scores of $x$'s above the optimal threshold.

### 2.3. An Asymptotic Equivalence Result

We now investigate the connections between EUM-optimal classifiers and DTA-optimal classifiers when the true distribution $P(X,Y)$ is known. By definition, a DTA-optimal classifier is expected to be better than an EUM-optimal classifier if tested on many i.i.d. test sets. We shall give an asymptotic equivalence result for EUM-optimal classifiers and DTA-optimal classifiers on large i.i.d test sets. In light of Theorem 4, we only need to consider an optimal probability-thresholding classifier as a representative EUM-optimal classifier.

In the following, let $\mathbf{x} = (x_1, \ldots, x_n) \in X^n$ be an i.i.d. sequence of observations. For any classifier $\theta$, let $\theta(\mathbf{x}) = (\theta(x_i))_i$. All expectations, denoted by $\mathrm{E}(\cdot)$, are taken under the conditional distribution $P(\mathbf{y}|\mathbf{x})$. The following theorem says that for an arbitrary classifier $\theta$, when $n$ is large enough, then for any $\mathbf{x}$, the expected F-measure of $\theta(\mathbf{x})$ is close to $F_\beta(\theta)$.

---

[2] Any $\theta \in \mathcal{T}$ can be approximated by members in $\mathcal{T}'$ with arbitrary close $F_\beta$, and vice versa, but $\mathcal{T}'$ may contain $\theta'$ such that $F_\beta(\theta) \neq F_\beta(\theta')$ for all $\theta \in \mathcal{T}$, implying $\mathcal{T}' \neq \mathcal{T}$.

**Theorem 6.** *For any classifier $\theta$, any $\epsilon, \eta > 0$, there exists $N_{\beta,\epsilon,\eta}$ such that for all $n > N_{\beta,\epsilon,\eta}$, with probability at least $1 - \eta$, $|\mathrm{E}[F_\beta(\theta(\mathbf{x}), \mathbf{y})] - F_\beta(\theta)| < \epsilon$.*

Such approximation holds uniformly for the class $\mathcal{T}$.[3]

**Lemma 7.** *For any $\epsilon, \eta > 0$, there exists $N_{\beta,\epsilon,\eta}$ such that for all $n > N_{\beta,\epsilon,\eta}$, with probability at least $1 - \eta$, for all $\delta \in [0, 1]$, $|\mathrm{E}[F_\beta(\mathrm{I}_\delta(\mathbf{x}), \mathbf{y})] - F_\beta(\mathrm{I}_\delta)| < \epsilon$.*

The above uniform approximation result leads to the following asymptotic equivalence result.

**Theorem 8.** *Let $\mathbf{s}^*(\mathbf{x}) = \max_\mathbf{s} \mathrm{E}[F_\beta(\mathbf{s}, \mathbf{y})]$, with $\mathbf{s}$ satisfying $\{P(1|x_i) \mid s_i = 1\} \cap \{P(1|x_i) \mid s_i = 0\} = \emptyset$. Let $t^* = \arg\max_{t \in \mathcal{T}} F_\beta(t)$. Then for any $\epsilon, \eta > 0$,*
*(a) There exists $N_{\beta,\epsilon,\eta}$ such that for all $n > N_{\beta,\epsilon,\eta}$, with probability at least $1 - \eta$, $\mathrm{E}[F_\beta(t^*(\mathbf{x}), \mathbf{y})] \leq \mathrm{E}(F_\beta(\mathbf{s}^*(\mathbf{x}), \mathbf{y})) < \mathrm{E}[F_\beta(t^*(\mathbf{x}), \mathbf{y})] + \epsilon$.*
*(b) There exists $N_{\beta,\epsilon,\eta}$ such that for all $n > N_{\beta,\epsilon,\eta}$, with probability at least $1 - \eta$, $|F_\beta(t^*(\mathbf{x}), \mathbf{y}) - F_\beta(\mathbf{s}^*(\mathbf{x}), \mathbf{y}))| < \epsilon$.*

Part (a) says that the $t^*(\mathbf{x})$ and $\mathbf{s}^*(\mathbf{x})$ have almost the same expected $F_\beta$, and Part (b) says that for a large i.i.d. test set $(\mathbf{x}, \mathbf{y})$, $t^*$ and $\mathbf{s}^*$ have almost identical $F_\beta$.

The constraint on $\mathbf{s}$ ensures that instances with the same probability of being positive are placed in the same class. In general, optimal predictions may not satisfy this constraint (Lewis, 1995). However, if the underlying distribution satisfies that $P(P(1|X) = \delta) = 0$ for any $\delta$, then the above result is essentially this: given $P$, an optimal prediction and the prediction using the optimal threshold are asymptotically equivalent. This is demonstrated empirically in Section 4.

## 3. Algorithms

We first discuss approximations to EUM, then discuss DTA and present a new efficient prediction algorithm.

### 3.1. Approximations to the EUM Approach

Exact empirical optimization of F-measures for a parametric family is difficult due to its complex piecewise linear nature, and typically only approximations of the F-measures are maximized. We discuss three methods.

In view of the optimality of probability thresholding classifiers, it is natural to first learn an estimate $p(Y|X)$ for $P(Y|X)$, and then learn an optimal threshold $\delta$. If $p(Y|X)$ is chosen from a parametric family using the maximum likelihood (ML) principle, then under very general conditions, the learned distribution follows an asymptotically normal convergence to the model with smallest KL-divergence to the true distribution (White, 1982). Thus when the model family is well-specified, the resulting classifier is asymptotically optimal. We call this the $ML^\delta$ *approximation*. Strictly, this is a combination of the conditional probability estimation and F-measure optimization of the threshold, and the convergence rate in Theorem 3 does not apply.

---
[3]Both Lemma 7 and Theorem 8 hold for $\mathcal{T} \cup \mathcal{T}'$ as well. We consider $\mathcal{T}$ to simplify the presentation.

Jansche (2005) learned a logistic regression model $p(Y|X, \phi)$ by maximizing the empirical $F_\beta$ in eq. 1, but with each binary decision $s_i$ replaced by the predictive probabilities $p_i = p(1|x_i, \phi)$. The eventual classifier uses the rule $h(x) = \mathrm{I}(p(1|x) > 0.5)$. It is unknown whether this method is consistent or whether it follows any asymptotic convergence. There is also no apparent reason to use 0.5 as the threshold, so we shall optimize the threshold on the training data in addition to estimating $\phi$. We call this the $F^\delta$ *approximation*.

We considered learning a rule $h(x) = \mathrm{I}(p(1|x, \phi) > \delta)$ directly, where $\phi, \delta$ are parameters, by approximating the empirical $F_\beta$ in eq. 1 using $s_i = \mathrm{I}^\gamma(p(1|x_i, \phi) - \delta)$, where $\mathrm{I}^\gamma(t) = 1/(1 + e^{-\gamma t})$ approximates $\mathrm{I}(t > 0)$ for large $\gamma$. However, this seemed to overfit easily, and it rarely yielded better performance than the $ML^\delta$ and $F^\delta$ approximations in our preliminary experiments. We will not consider it further.

### 3.2. Maximizing Expected F-measure

Given a utility function $U(\mathbf{s}, \mathbf{y})$, the decision-theoretic optimal prediction for $\mathbf{x}$ maximizes $\mathrm{E}_{\mathbf{y} \sim P(\cdot|\mathbf{x})}(U(\mathbf{s}, \mathbf{y}))$. In general, the true distribution $P$ is not known and is estimated. The approach that involves first estimating true distributions using maximum likelihood (ML) and then making decision-theoretic optimal predictions will be called the $ML^E$ *approach*. We discuss the two steps in $ML^E$, then we present an efficient algorithm for computing the optimal predictions.

First, the asymptotic convergence of ML (White, 1982) implies the $ML^E$ approach is asymptotically optimal when estimating with sufficient training examples in a well-specified family. In practice, we will not know whether the model family is well-specified. Nevertheless, as we shall see in Section 4.1, the $ML^E$ approach can yield results indistinguishable from the optimal if the model family is misspecified but contains a reasonable approximation to the true distribution.

Second, for arbitrary utility function $U$, computing the expectation can be computationally difficult. But for the case when the utility function is an F-measure, and

**Algorithm 1** Compute $f_{\beta;1}, \ldots, f_{\beta;n}$, where $\beta^2 = q/r$

1: For $0 \leq i \leq n$, set $C[i]$ as the coefficient of $x^i$ in $[p_1 x + (1-p_1)] \ldots [p_N x + (1-p_N)]$;
2: For $1 \leq i \leq (q+r)n$, $S[i] \leftarrow q/i$;
3: **for** $k = n$ to $1$ **do**
4: $\quad f_{\beta;k} \leftarrow \sum_{k_1=0}^{n} (1+r/q) k_1 C[k_1] S[rk + qk_1]$;
5: $\quad$ Divide $C$ by $p_k x + (1-p_k)$;
6: $\quad$ **for** $i = 1$ to $(q+r)(k-1)$ **do**
7: $\quad\quad S[i] \leftarrow (1-p_k) S[i] + p_k S[i+q]$;
8: $\quad$ **end for**
9: **end for**

$P(\mathbf{y}|\mathbf{x}) = \prod_{i=1}^{n} P(y_i|x_i)$, efficient algorithms can be designed by exploiting the following characterization of an optimal prediction. Let $p_i = P(1|x_i)$.

**Theorem 9.** *(Probability Ranking Principle for F-measure, Lewis 1995)* Suppose $\mathbf{s}^* = \max_{\mathbf{s}} \mathrm{E}(F_\beta(\mathbf{s}, \mathbf{y}))$. Then $\min\{p_i \mid s_i^* = 1\} \geq \max\{p_i \mid s_i^* = 0\}$.

Thus the decision-theoretic optimal prediction contains the top $k$ instances that are most likely to be positive for some $k \in \{0, \ldots, n\}$. This reduces the number of candidate predictions from $2^n$ to $n+1$. We shall use this result to give an efficient algorithm for computing the optimal predictions.

#### 3.2.1. A Quadratic Time Algorithm

We give an $O(n^3)$ time algorithm for computing the optimal predictions, then improve it to $O(n^2)$ when $\beta^2$ is rational, which is often the case.

Let $F_{\beta;k}(\mathbf{y})$ be the $F_\beta$-measure when the first $k$ instances are predicted as positive, then we have $F_{\beta;k}(\mathbf{y}) = (1+\beta^2) \sum_{i=1}^{k} y_i / [k + \beta^2 \sum_{i=1}^{n} y_i]$. Let $f_{\beta;k} = \sum_{\mathbf{y}} P(\mathbf{y}) F_{\beta;k}(\mathbf{y})$, and $S_{i:j} = \sum_{l=i}^{j} y_l$. For $\mathbf{y}$'s satisfying $S_{1:k} = k_1$ and $S_{k+1:n} = k_2$, their $F_\beta$'s are $(1+\beta^2)k_1/(k + \beta^2(k_1 + k_2))$, and the probability this happens is $P(S_{1:k} = k_1) P(S_{k+1:n} = k_2)$, thus

$$f_{\beta;k} = \sum_{\substack{0 \leq k_1 \leq k \\ 0 \leq k_2 \leq n-k}} \frac{P(S_{1:k} = k_1) P(S_{k+1:n} = k_2)(1+\beta^2) k_1}{k + \beta^2(k_1 + k_2)}.$$

One can show that $P(S_{1:k} = i)$ and $P(S_{k+1:n} = i)$ are the coefficients of $x^i$ in $\prod_{j=1}^{k}[p_j x + (1-p_j)]$ and $\prod_{j=k+1}^{n}[p_j x + (1-p_j)]$ respectively. Thus, each $f_{\beta;k}$ can be computed in $O(n^2)$ time using $O(n)$ space. Hence computing all $f_{\beta;k}$'s takes $O(n^3)$ time and $O(n)$ space.

For rational $\beta^2$, we can improve the computation to $O(n^2)$ time and $O(n)$ space. The key is to note that

$$f_{\beta;k} = \sum_{k_1=0}^{k} (1+\beta^{-2}) k_1 P(S_{1:n} = k_1) s(k, k\beta^{-2} + k_1),$$

where $s(k, \alpha) = \sum_{k_2=0}^{n-k} P(S_{k+1:n} = k_2)/(\alpha + k_2)$. For rational $\beta$, the $s$ values required for the $f_{\beta;k}$'s are shared. To compute $s$, use $s(n, \alpha) = 1/\alpha$, and

$$s(k-1, \alpha) = p_k s(k, \alpha+1) + (1-p_k) s(k, \alpha),$$

which follows from $P(S_{k:n} = i) = p_k P(S_{k+1:n} = i - 1) + (1-p_k) P(S_{k+1:n} = i)$.

The pseudo-code is given in Algorithm 1, with $q/r$ as the reduced fraction of $\beta^2$. Correctness can be seen by observing that at line 3, $S[i] = s(k, i/q)$, and $C[k_1] = P(S_{1:k} = k_1)$. In practice, polynomial division can be numerically unstable, and it is preferred to precompute all the $C[i]$'s using $O(n^2)$ time and space first.

### 4. Experiments

We empirically demonstrate that EUM can be more robust against model misspecification, but DTA can be better for rare classes on small datasets and a common scenario of domain-adaptation. We use a synthetic dataset, the Reuters-21578 dataset, and four multi-label classification datasets.

#### 4.1. Mixtures of Gaussians

We consider a mixture of Gaussians on $D$ dimensions: $P(X, Y) = \pi_Y N(X; \mu_Y, \Sigma_Y)$, with $\Sigma_1 = \Sigma_0 = I_D$, $\mu_1 = (S+O)\mathbf{1}/\sqrt{4D}$ and $\mu_0 = -(S-O)\mathbf{1}/\sqrt{4D}$, where $S$ and $O$ are non-negative constants. Thus $S$ is the distance between the centers. We shall vary $S$, $O$, $D$, $\pi_1$ and the number of training examples $N_{tr}$. All instances are i.i.d. The optimal $F_1$ achievable by a classifier $\theta$ can be computed (see eq. 2), and it depends only on $S$ and $\pi_1$. $N_{tr}$ determines how close the estimated distribution is to the optimal model; and the number of test examples, $N_{ts}$, affects the gap in the performance between the thresholding method and the expectation method (Theorem 8).

We train logistic regression (LR) models using three different attribute vector representations: $R_0$ consists of the coordinates only, $R_1$ is $R_0$ with an additional dummy attribute fixed at 1, and $R_2$ is $R_1$ with additional all degree two monomials of the coordinates. LR with $R_2$ includes the true distribution. The methods compared are $\mathrm{ML}^\delta$, $\mathrm{F}^\delta$, $\mathrm{ML}^E$, $\mathrm{Truth}^\delta$ and $\mathrm{Truth}^E$, where last two methods use the true model $P(X, Y)$ for thresholding and expectation.

The first column in Table 1 lists the parameter settings. For the row headed by *Default*, we use $D = 10$, $S = 4$, $O = 0$, $N_{tr} = 1000$, $N_{ts} = 3000$, and $\pi_1 = 0.5$. This dataset is low dimensional, almost noiseless, balanced and has sufficiently many train and test in-

**Optimizing F-Measures: A Tale of Two Approaches**

Table 1. Performance of different methods for optimizing $F_1$ on mixtures of Gaussians

| Setting | $\mathrm{ML}^E$ | | | $\mathrm{ML}^\delta$ | | | $\mathrm{F}^\delta$ | | | $\mathrm{Truth}^E$ | $\mathrm{Truth}^\delta$ | Theory |
| --- | --- | --- | --- | --- | --- | --- | --- | --- | --- | --- | --- | --- |
| | $R_0$ | $R_1$ | $R_2$ | $R_0$ | $R_1$ | $R_2$ | $R_0$ | $R_1$ | $R_2$ | | | |
| Default | 97.87 | 97.84 | 96.02 | 97.84 | 97.87 | 96.15 | 97.62 | 97.55 | 96.37 | 97.87 | 97.91 | 97.72 |
| $S=0.4$ | 66.86 | 66.86 | 63.77 | 66.32 | 66.31 | 63.55 | 66.03 | 66.09 | 65.72 | 66.39 | 65.82 | 66.88 |
| $D=100$ | 94.12 | 94.14 | 88.05 | 94.09 | 94.08 | 87.86 | 95.96 | 95.98 | 88.23 | 97.53 | 97.53 | 97.72 |
| $N_{tr}=100$ | 95.43 | 95.48 | 91.36 | 94.78 | 94.69 | 91.33 | 95.55 | 95.34 | 91.57 | 97.80 | 97.36 | 97.72 |
| $\pi_1=0.05$ | 75.19 | 90.79 | 84.07 | 91.84 | 90.17 | 84.21 | 92.36 | 89.56 | 85.21 | 92.72 | 92.26 | 91.73 |
| $O=50$ | 66.01 | 67.83 | 96.10 | 65.44 | 89.29 | 96.10 | 97.04 | 96.88 | 97.41 | 97.87 | 97.91 | 97.72 |

stances.[4] Each of the remaining rows uses the same set of parameters, except for the one parameter indicated on the first column. LR with $R_0$ or $R_1$ contains a good approximation to the true distribution for all settings except $\pi_1 = 0.05$ and $O = 50$. For $\pi_1 = 0.05$, the class is imbalanced and such imbalance cannot be modelled without the dummy attribute. Thus $R_0$ will not give a good model, but $R_1$ will. For $O = 50$, the centers are far from the origin, and this makes both $R_0$ and $R_1$ inadequate for density estimation.

In Table 1, the $F_1$ results for $\mathrm{Truth}^E$, $\mathrm{Truth}^\delta$ and Theory (the theoretical optimal $F_1$) are similar. These are expected according to Theorem 8. Most other scores are close to the optimal scores. For $\mathrm{ML}^E$ and $\mathrm{ML}^\delta$, these scores are expected due to the presence of a good approximation to the true distribution in the model family, and the asymptotic convergence property of $\mathrm{ML}^\delta$ and $\mathrm{ML}^E$ given sufficiently many examples, as discussed in Section 3. For $\mathrm{F}^\delta$, although we lack its theoretical convergence to an optimal classifier, the results suggest that such convergence may hold.

The scores obtained using $R_2$ are generally lower than scores obtained using $R_0$ and $R_1$ under the settings *Default*, $S = 0.4$, $D = 100$, and $N_{tr} = 100$, though $R_2$ gives a well-specified model class while $R_1$ and $R_0$ do not. Thus, a well-specified model class is not necessarily better. This is because a misspecified model class with a small VC dimension can converge to the optimal model within the class using fewer samples than a well-specified model class with a higher VC dimension. To choose a class of the right complexity, one may follow the structural risk minimization principle (Vapnik, 1995). This requires bounds like those in Lemma 2 and Theorem 3. However, the given bounds cannot be used because they only apply for large samples.

The gaps between $R_2$ scores and the optimal score for *Default* is significantly smaller than the gaps for $S =$

---

[4]We have verified that the sizes are large enough to give the same conclusions for other i.i.d. data of the same sizes.

$0.4$, $D = 100$, $N_{tr} = 100$, and $\pi_1 = 0.05$. This suggests that higher noise level, higher model class complexity, smaller training size, and smaller positive ratio make it harder to learn a good classifier. Note that Theorem 3 already suggests that in EUM, smaller positive ratio can make learning more difficult.

For the setting $\pi_1 = 0.05$, using $R_0$, $\mathrm{ML}^E$ performs poorly, while $\mathrm{ML}^\delta$ is close to optimal. $\mathrm{ML}^E$'s poor performance is expected due to poor quality of the learned distribution, and $\mathrm{ML}^\delta$'s performance can be justified by Theorem 5: the thresholding method can remain optimal when the score function is rank-preserving but not close to the true probability distribution. For the setting $O = 50$, both $\mathrm{ML}^E$ and $\mathrm{ML}^\delta$ perform poorly using $R_0$, but $\mathrm{ML}^\delta$ is much better than $\mathrm{ML}^E$ using $R_1$. Thus although $\mathrm{ML}^\delta$ can still be severely affected by model misspecification, it is still relatively robust. In addition, for $\pi_1 = 0.05$ and $O = 50$, $\mathrm{F}^\delta$ has much higher or at least comparable scores than $\mathrm{ML}^E$ and $\mathrm{ML}^\delta$. This suggests that *if the model class is severely misspecified, then EUM can be more robust than DTA*.

We also compare $\mathrm{ML}^E$ and $\mathrm{ML}^\delta$ on small test sets with $N_{ts} = 100$ (Theorem 8 only holds for large test set size). We observed similar performances from $\mathrm{ML}^\delta$ and $\mathrm{ML}^E$ when $\pi_1$ is high, but $\mathrm{ML}^E$ seems significantly better than $\mathrm{ML}^\delta$ when $\pi_1$ is small. To illustrate, Table 2 gives the results when the same setting as $\pi_1 = 0.05$ in Table 1 is used to generate the data. It shows that, *with a sufficiently accurate model, $\mathrm{ML}^E$ can be better than $\mathrm{ML}^\delta$ and $\mathrm{F}^\delta$ on rare classes*.

#### 4.1.1. EFFECT OF MODEL QUALITY

We also perform experiments to study the effect of incorrect probability models on $\mathrm{ML}^E$. We use the *Default* setting in the previous section, with $\pi_1 = 0.5$ and $S = 4$ changed to $S = 2$, as the true distribution, to generate a set of 3000 i.i.d. test instances. We make optimal predictions using an assumed distribution which is the same as the true one except that we vary $\pi_1$. For each $\pi_1$, we compute the $F_1$ and the Kullback-

Optimizing F-Measures: A Tale of Two Approaches

Table 2. The means and standard deviations of the $F_1$ scores in percentage, computed using 2000 i.i.d. trials, each with test set of size 100, for mixtures of Gaussians with $D = 10$, $S = 4$, $O = 0$, $N_{tr} = 1000$ and $\pi_1 = 0.05$.

|  | $ML^E$ | | | $ML^\delta$ | | | $F^\delta$ | | | $Truth^E$ | $Truth^\delta$ |
| --- | --- | --- | --- | --- | --- | --- | --- | --- | --- | --- | --- |
|  | $R_0$ | $R_1$ | $R_2$ | $R_0$ | $R_1$ | $R_2$ | $R_0$ | $R_1$ | $R_2$ | | |
| Mean | 36.70 | 63.00 | 58.80 | 60.78 | 62.01 | 58.31 | 61.40 | 59.98 | 53.16 | 63.32 | 60.71 |
| Std. dev. | 13.04 | 20.67 | 21.49 | 23.69 | 21.34 | 21.87 | 22.18 | 22.04 | 23.02 | 20.46 | 23.72 |

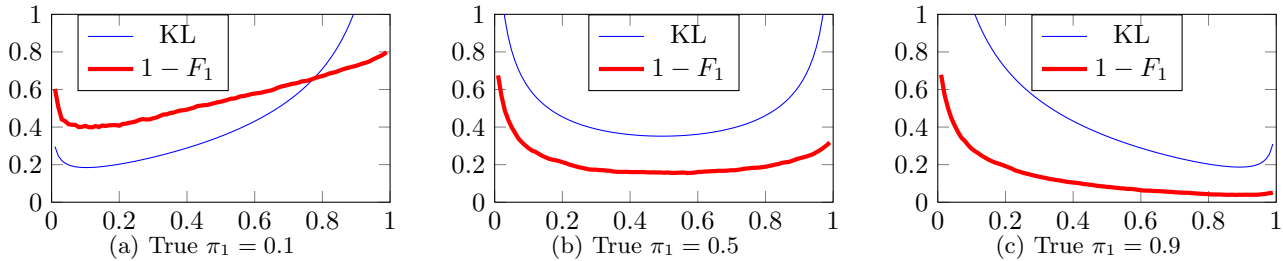

Figure 1. Effect of the quality of probability model on the decision-theoretic method. The $x$-axes are the $\pi_1$ values for the assumed distribution, and the $y$-axes are the corresponding $1 - F_1$ and KL values.

Leibler-divergence (KL) from the true to the assumed distribution on the test set. These are plotted in Figure 1(b), where $1 - F_1$ is plotted instead of $F_1$. Figures 1(a) and 1(c) plot for similar experiments, but using 0.1 and 0.9 as the true $\pi_1$ instead. Our choice of $S = 2$ instead of $S = 4$ for the true distribution has made the difference between the true and assumed distributions more pronounced in the plots. Comparing the curves for KL and $1 - F_1$ within each figure, we see that the F-measure of DTA is roughly positively correlated with the model quality. The plot for $1 - F_1$ in Figure 1(a) exhibits higher curvature around the true $\pi_1$ than those in the other two figures. This suggests that if the true distribution has a small positive ratio, the performance is more sensitive to model quality.

4.1.2. DOMAIN ADAPTATION

In domain adaptation, the test distribution differs from the training one. One common scenario is when $P(X)$ changes but $P(Y|X)$ does not. Using the mixture of Gaussians with $D = 10$, $S = 4$, $O = 0$ and $\pi_1 = 0.5$, we generate 5000 i.i.d. training instances, and 5000 test instances with $P(Y|X) < 0.5$. The $F_1$ scores for $Truth^\delta$, $Truth^E$, $ML^\delta$ and $ML^E$ (using $R_1$) are 21%, 38%, 11% and 36% respectively. Similar results are obtained under similar settings. Under such conditions, DTA is more robust than EUM.

**4.2. Text Classification**

We evaluate on the Reuters-21578 dataset[5] using the ModApte partition, which has 9603 training documents and 3299 test documents. We train two models: the standard multinomial naïve Bayes (NB) model and a LR model, using word occurrence counts and a dummy attribute fixed at one. Both models are regularized. For NB, we use the Laplace corrector with one count for class and word counts. For LR, we use the Gaussian norm on the parameters. We use only those topics with at least $C$ positive instances in both the train and test sets, and we vary $C$. Table 3 reports macro-$F_1$ scores (the $F_1$ averaged over topics), where $ML^{.5}$ uses 0.5 to threshold the probabilities,

Table 3. Macro-$F_1$ scores in percentage on the Reuters-21578 dataset, computed for those topics with at least $C$ positive instances in both the training and test sets. The number of topics down the rows are 90, 50, 10 and 7.

|  | Naïve Bayes | | | Logistic regression | | | |
| --- | --- | --- | --- | --- | --- | --- | --- |
| $C$ | $ML^{.5}$ | $ML^\delta$ | $ML^E$ | $ML^{.5}$ | $ML^\delta$ | $F^\delta$ | $ML^E$ |
| 1 | 17.4 | 17.7 | 17.7 | 35.8 | 36.6 | 37.3 | 39.9 |
| 10 | 29.8 | 30.1 | 30.2 | 55.1 | 56.4 | 57.2 | 57.6 |
| 50 | 69.9 | 69.1 | 70.1 | 75.2 | 75.7 | 76.6 | 75.6 |
| 100 | 73.7 | 73.5 | 73.7 | 75.5 | 75.9 | 76.5 | 75.8 |

In Table 3, although NB generally does not provide good probability estimates, $ML^E$ is still at least comparable to $ML^{.5}$ and $ML^\delta$. With LR, $ML^E$ is a few percents better for rare classes. Chai (2005) used Gaussian process and obtained similar conclusion.

**4.3. Multilabel Datasets**

We evaluate on four standard multilabel classification datasets.[6] We train regularized LR, with the regu-larization parameter for each class selected using two fold cross validation. Macro-$F_1$ scores are shown in Table 4. The bracketed scores are obtained by choosing the regularization parameter giving a model with minimum empirical KL divergence on the test data. Each bracketed score is higher than its non-bracketed counterpart, thus models closer to the true one perform better for both $\mathrm{ML}^E$ and $\mathrm{ML}^\delta$. Comparing the scores for $\mathrm{ML}^E$ with those for $\mathrm{ML}^\delta$ and $\mathrm{F}^\delta$, bracketed or not, we see that $\mathrm{ML}^E$ performs better, especially for smaller $C$, suggesting $\mathrm{ML}^E$ is better for rare classes.

---
[5]This is from http://www.daviddlewis.com/resources/testcollections/reuters21578/.

[6]These are available at http://mulan.sourceforge.net/.

Table 4. Macro-$F_1$ scores in percentage on four multilabel datasets, computed for those $T$ labels with at least $C$ positive instances in both the training and test sets.

| $C$ | $T$ | $\mathrm{ML}^\delta$ | $\mathrm{ML}^E$ | $\mathrm{F}^\delta$ |
|---|---|---|---|---|
| **yeast (1500 train, 917 test)** | | | | |
| 1 | 14 | 47.14 (47.54) | 48.16 (48.47) | 46.61 |
| 50 | 13 | 50.76 (50.34) | 51.38 (51.71) | 50.19 |
| 300 | 5 | 73.79 (73.31) | 73.52 (73.74) | 73.71 |
| **medical (645 train, 333 test)** | | | | |
| 1 | 32 | 48.88 (51.93) | 51.48 (53.91) | 48.45 |
| 10 | 12 | 84.81 (84.49) | 85.19 (85.84) | 86.01 |
| 50 | 2 | 87.62 (88.78) | 90.12 (88.99) | 91.56 |
| **scene (1211 train, 1196 test)** | | | | |
| 1 | 6 | 68.80 (70.50) | 68.57 (70.80) | 69.05 |
| 100 | 6 | 68.80 (70.50) | 68.57 (70.80) | 69.05 |
| **enron (1123 train, 579 test)** | | | | |
| 1 | 52 | 19.70 (25.53) | 21.61 (26.45) | 19.24 |
| 10 | 26 | 35.26 (38.00) | 38.76 (39.74) | 35.86 |
| 50 | 9 | 59.21 (59.82) | 60.15 (60.44) | 61.60 |

## 5. Conclusion

We gave theoretical justifications and connections for optimizing F-measures using EUM and DTA. We empirically demonstrated that EUM seems more robust against model misspecification, while given a good model, DTA seems better for handling rare classes and a common domain adaptation scenario.

A few important questions are unanswered yet: existence of interesting classifiers for which EUM can be done exactly, quantifying the effect of inaccurate models on optimal predictions, identifying conditions under which one method is preferable to another, and practical methods for selecting the best method on a dataset. Results in this paper only hold for large data sets, and it is important to consider the case for small number of instances. Experiments with and analyses of other methods may yield additional insights as well.

## Acknowledgement

This work is supported by DSO grant DSOL11102.